\documentclass[runningheads]{llncs}
\usepackage{graphicx}
\usepackage{hyperref}
\usepackage{color}
\usepackage{ragged2e}
\usepackage{float}

\usepackage{comment}

\usepackage{array}
\usepackage{multirow}

\newcommand\MyBox[2]{
  \fbox{\lower0.75cm
    \vbox to 1.7cm{\vfil
      \hbox to 1.7cm{\hfil\parbox{1.4cm}{#1\\#2}\hfil}
      \vfil}%
  }%
}

\usepackage{xcolor}
\usepackage[linesnumbered,ruled,vlined]{algorithm2e}

\SetCommentSty{mycommfont}

\SetKwInput{KwInput}{Input}                
\SetKwInput{KwOutput}{Output}              

\usepackage{amssymb}
\usepackage{amsmath}
\usepackage{tikz}
\usetikzlibrary{positioning}
\usepackage{geometry}
\usepackage{tabularx}
\usepackage{paracol}
\begin{document}
\title{Bi-Attention HateXplain : Taking into account the sequential aspect of data during explainability in a multi-task context}
%
\titlerunning{Bi-Attention HateXplain}
%
\author{Ghislain Dorian Tchuente Mondjo \inst{1} \\
\email{tchuente.mondjo@gmail.com}
}
\authorrunning{G. Tchuente}
%
\institute{Université de Yaoundé I, Faculté des Sciences, Département d’Informatique \\ BP 812 Yaoundé, Cameroun 
}
\maketitle              
\begin{abstract}
Technological advances in the Internet and online social networks have brought many benefits to humanity. At the same time, this growth has led to an increase in hate speech, the main global threat. To improve the reliability of black-box models used for hate speech detection, post-hoc approaches such as LIME, SHAP, and LRP provide the explanation after training the classification model. In contrast, multi-task approaches based on the HateXplain benchmark learn to explain and classify simultaneously. However, results from HateXplain-based algorithms show that predicted attention varies considerably when it should be constant. This attention variability can lead to inconsistent interpretations, instability of predictions, and learning difficulties. To solve this problem, we propose the BiAtt-BiRNN-HateXplain (Bidirectional Attention BiRNN HateXplain) model which is easier to explain compared to LLMs which are more complex in view of the need for transparency, and will take into account the sequential aspect of the input data during explainability thanks to a BiRNN layer. Thus, if the explanation is correctly estimated, thanks to multi-task learning (explainability and classification task), the model could classify better and commit fewer unintentional bias errors related to communities. The experimental results on HateXplain data show a clear improvement in detection performance, explainability and a reduction in unintentional bias.
 \\

\textbf{Warning :} This article contains material that many will find offensive or hateful; however, this is unavoidable given the nature of the work.
\footnote{Accepted at ``EAI AFRICOMM 2025 - \textit{17th EAI International Conference \\on Communications and Networks in Africa}''. Copyright \textcopyright{} 2025 EAI.}
\end{abstract}
\begin{multicols}{2}
\section{Introduction}The proliferation of hate speech online is a serious threat capable of compromising security, social cohesion, and even living together. To counter this problem, a considerable amount of research has focused on detecting hate speech \cite{3},\cite{4},\cite{5} leading to a wide variety of recognition approaches. Many of them leverage the power of artificial intelligence (AI) and machine learning (ML) techniques to automatically extract and learn features from the data itself. Despite current limitations in accuracy, ML-based approaches can significantly reduce the amount of manual effort needed to monitor online platforms. At the same time, recent methods have become opaque and very complex for humans \cite{6}. Deep Learning models have gained popularity due to their success in practice, but behave like black boxes \cite{6}. A new field of research, called eXplainable Artificial Intelligence (XAI), has emerged that seeks to explain the behavior of complex and opaque models. Whether increasing social acceptance or ensuring fairness in decision-making, machine learning research would benefit enormously from interpretable model predictions \cite{7}.
The recognition of hate speech is no exception here. There is a clear need to explain the models' decisions to classify speech as hateful or not. This becomes even more obvious if their predictions are used to implement countermeasures against supposedly hateful users. Several approaches to explainability of Deep Learning models have been proposed. Post-hoc methods will provide explainability from the classification model through several technical approaches. These encompass \textit{perturbation-based methods}, which assume that eliminating crucial hate speech components significantly impacts classification results; \textit{local substitution methods} that construct transparent local surrogates, exemplified by LIME's linear models \cite{8}; and \textit{}{backpropagation methods} that determine feature relevance efficiently via single-pass propagation, as demonstrated by LRP \cite{10} and supported by \cite{9}.

However, the drawback of these two previous approaches is that their \textit{predicted attention varies when it should be constant}, which can lead to inconsistent interpretation, instability of predictions, and learning difficulties. We propose to combine the idea of Xu in \cite{19} that uses a layer compatible with the structure of the input data to estimate the explanation and the idea of Mathew in \cite{1} that proposes to take into account the dependency between explainability and classification using multi-task learning. 
In the context of hate speech detection, where social, ethical, and legal issues are major, the explainability of models is an essential requirement. Although LLM (Large Language Models) models such as GPT or BERT offer impressive performance in text classification, their structural complexity, massive number of parameters, and opaque operation make their behavior difficult to interpret. Conversely, more compact architectures such as BiRNNs (Bidirectional Recurrent Neural Networks) allow for better traceability of decisions made, in particular thanks to their ability to process text word by word in both temporal directions. Work such as that of Wang and al. from 2019 to 2022 on State-Regularized RNNs or of Merrill \& Tsivilis in 2022 has shown that it is possible to extract deterministic finite automata (DFA) from RNNs, allowing visualization of internal states and transitions. Thus, with the aim of transparency and accountability, particularly towards users or moderators, BiRNNs appear to be a relevant and more interpretable alternative to LLMs, while remaining sufficiently efficient on targeted and well-annotated datasets.
We propose to take into account the sequential aspect of the data when estimating attention or explanation and propose the BiAtt-BiRNN-HateXplain algorithm, inspired by the BiRNN-HateXplain algorithm \cite{1}, which, instead of using the attention mechanism proposed in \cite{11}, instead uses a BiRNN layer for explainability. BiRNN \cite{12} was chosen because it is well suited to managing the sequential aspect of a sentence. The BiRNN makes it possible to extract the precedence features by processing the sequence from left to right and from right to left, which will allow having the context representations on the left and right of each word and therefore have the overall context of each word in a sentence. The rest of our work will be divided as follows: In section 1, we will talk about the state of the art of explainability and the detection of hate speech in artificial intelligence; in section 2 we will present the limits of current multitask explainability models, the idea to overcome this difficulty and then the BiAtt-BiRNN-HateXplain algorithm. Finally, section 3 we will present the experiments and the limits of the proposed model.
\section{Related Work}
In recent times, with the advent of social media networks and other types of online platforms that incorporate user-generated content, online communities have witnessed a growing phenomenon of abusive and hateful behavior among their members. The effects can be devastating for those targeted: high levels of anxiety and stress and, in some cases, depression. This demonstrates that hostile behavior online is no longer just a sporadic occurrence. In fact, according to research from \cite{13}, 66\% of American adults have witnessed some form of online harassment. Given the prevalence of the phenomenon, there is no doubt that manual controls and prevention are by no means an effective solution. The urgent need for automated and scalable methods to combat online abuse has motivated intensive research in the area.
\subsection{Hate speech detection}
When we look at hate speech detection models, a state-of-the-art transition (SOTA) has taken place in recent years. We have moved from methods using manual feature extraction and hand-crafted rules \cite{14},\cite{22}, \cite{23} to approaches that involve more automation of the feature extraction process, such as deep neural network (DNN). The use of lexicon-based features only appeared 13 years later in \cite{15}. For any text classification task, it is quite common to use Bag of Words (BoW)(as of 2012) to represent the text. Although using $BoW$ features generally gives good classification performance, the effectiveness of this representation strongly depends on whether the predictive words appear in the training and testing set. More recently, with the widespread adoption of deep neural network (DNN) algorithms, a technique called words embedding has been proposed as an enhanced representation of words and sentences. Other works, starting with \cite{16}, have taken a different route and considered including social characteristics and personal traits alongside the text.

\begin{figure}[H]
    \centering
    \includegraphics[scale=0.4]{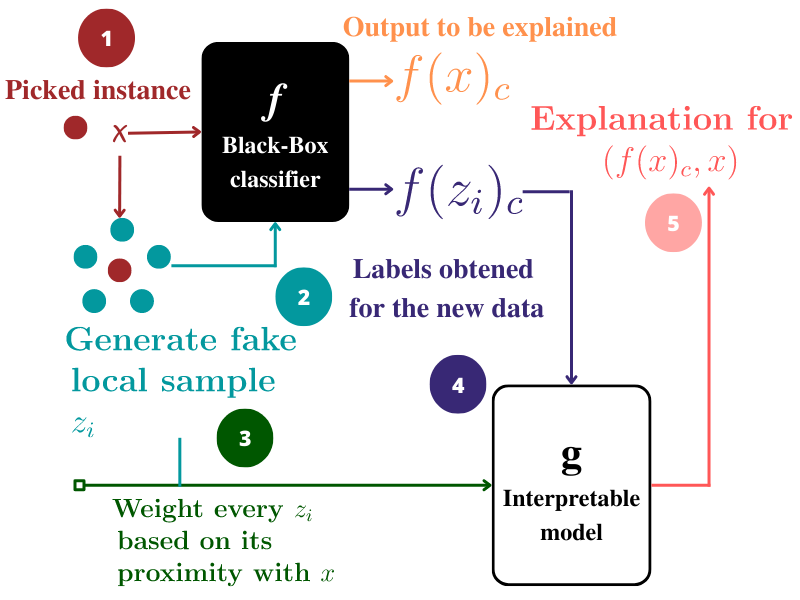}
    \caption{Flow of the LIME approach to explain the prediction of a classifier $f$ on an instance $x$.\cite{18}}
    \label{fig:12}
\end{figure}

\subsection{Explicabily of black box models}
According to IBM Watson, “Explainable artificial intelligence (XAI) is a set of processes and methods that enable human users to understand and trust the results and conclusions created by machine learning algorithms.” Explainable AI is crucial for an organization that wants to build trust when putting black-box AI models into production that are impossible to interpret. We distinguish three types of systems: Transparent models, Post-hoc models, Models based on multitask learning. Transparent models are inherently interpretable by design, but post-hoc models instead describe a framework in which a method of explanation targets the non-transparent model and extracts information. One of the main advantages of this concept is that it can be applied after training without sacrificing prediction performance \cite{2}. In the literature, different explainability models such as LIME, SHAP and LRP which locally explain a prediction model.

\paragraph{Local Interpretable Model-Agnostic Explanations (LIME)} Proposed by Ribeiro in \cite{8}, LIME is one of the explanation methods that directly implements the idea of a \textit{local surrogate}. It defines an interpretable model $g$ locally faithful to the classifier to be explained. Before examining the mathematical details of LIME, we will sketch the steps of the method without formality. We do this with the support of figure \ref{fig:12}. Choose a classifier, an input instance and the corresponding output that we want to explain. Generate points by perturbing the selected instance and, for each new point, obtain the classifier's prediction. Weight newly generated samples based on proximity to the originally selected instance. Choose an interpretable model and train it on the generated weighted variation dataset and labels obtained in step 2. Explain the original prediction by interpreting the simple local model.
\begin{figure}[H]
 \centering
 \includegraphics[width=9.3cm]{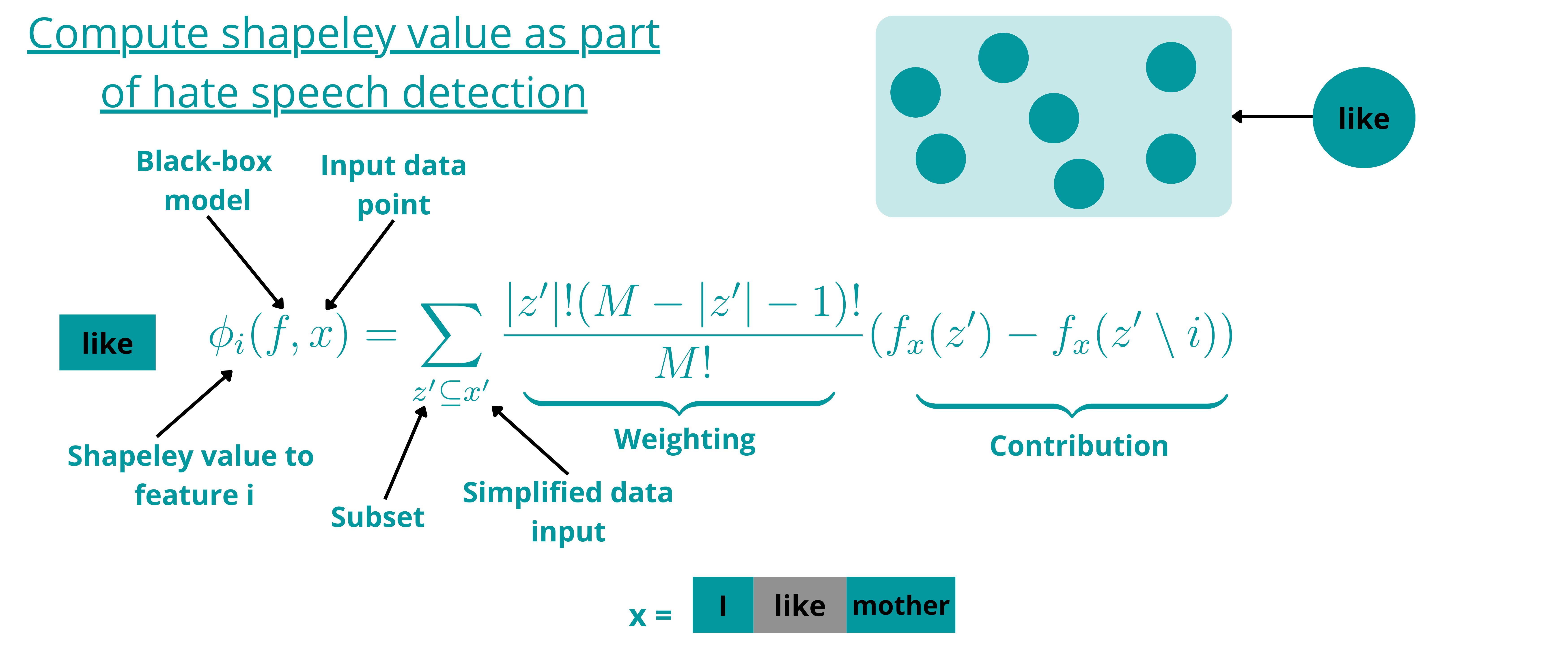}
 \caption{Study of the SHAP approach to explain the prediction of a classifier $f$ on an instance $x$.}
 \label{fig:23}
\end{figure}
\paragraph{SHapley Additive exPlanations (SHAP)}We also have SHAP \cite{18,20} which will explain in a model locally, by calculating the Shapley values of each characteristic from the impact of this characteristic in relation to the possible parts of the characteristic. 
Let $f$ be a black box model and $x$ be an example of the input data. Consider the output $f(x)_c$ to be explained and let $S$ be a subset of the characteristics of $x$, i.e. $S \subset {\{x_t\}}_{t=1}^M$. Let $x_S$ be the entry $x$ where entities not in $S$ are set to $0$. So, for $t = \{1, ..., M\}$, the Shapley value $\phi_t$ of $x_t$ is
{\scriptsize
\begin{equation}
   \phi_t = \sum_{S \subset {\{x_i\}}_{i=1}^M \setminus \{x_t\}} \frac{|S|!(M-|S|-1) !}{M!}(f(x_{S \cup \{x_t\}}) - f(x_{S})) 
\end{equation}}
and, for $t = 0$, we have the default $\phi_0 = f(x_{\emptyset})$. $\frac{|S|!(M-|S|-1)!}{M!}$ is simply the number of possible choices for the subset $S$ (it can be seen as a normalization term). Figure \ref{fig:23} presents an example of application of SHAP in the detection of hate speech. The objective is to calculate the relevance or contribution $\phi_i$ for each word of x = "\textit{I like mother}".
\paragraph{Layer-wise Relevance Propagation (LRP)}We finally have LRP \cite{9} for deep neural network models which will explain by calculating what is called \textit{relevance} iteratively, from the output class neurons to the first input neurons. We can think of relevance as a quantity that goes from output to input. The concept of relevance score for an input feature $x_t$ to a general unit $u$ in $f$ allows us to take advantage of the layered structure of $f$. Therefore, we denote $R_u^{(l)}$ as the relevance score of a unit $u$ in layer $l$. Using these quantities, we can decompose a target class output $f(x)_c$ into relevance by input feature $x_t$, or equivalently, by unit $u$ belonging to the first layer $1$. That's to say
\[ f(x)_c \approx \sum_{t=1}^{|x|} \phi(t, c, x) = \sum_{u \in 1} R_u^{(1)} \]
When backpropagating the relevance of a unit $u$ in layer $l$, we would expect to only backpropagate to those features in layer $l-1$ that contributed to $u$. Furthermore, the amount of relevance that backpropagates must be proportional to the contribution, that is, relate to the weight $w_{vu}$ for each $v \in (l - 1)$.
Then, LRP steps back one layer at a time and assigns relevance scores until the input layer is reached. As a rule for redistribution from one layer to the previous one, the following recursive formula is used:
\[ R_u^{(l)} = \sum_{v \in (l+1)} R_{v}^{(l+1)} \frac{a_u w_{uv}}{a_v + \varepsilon . sign(a_v)} \]
\noindent
Here, $a_u$ and $w_{uv}$ are used respectively to indicate the activation of $u$ and the weight between $u$ and $v$. But, the LIME algorithm is unstable, SHARP is time-consuming and LRP is not agnostic for all deep learning models.
\begin{figure}[H]
    \centering
    \includegraphics[width=7.5cm]{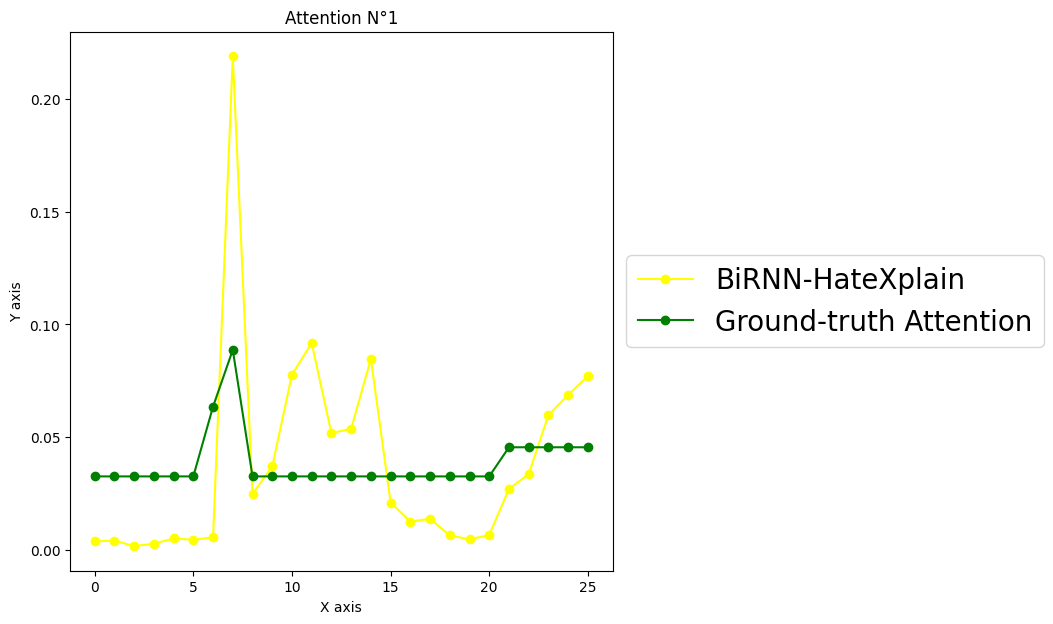}
    \caption{Attention predicted by BiRNN-HateXplain HateXplain in \textcolor{yellow}{yellow} and ground truth attention in \textcolor{green}{green}}
    \label{fig:1}
\end{figure} 
\paragraph{HateXplain Models based on multitask learning}Explainability models based on multitask learning of classification and explainability proposed in 2021 by \cite{1} are models based on the HateXplain benchmark for detecting hate speech which take into account the dependence between the task of explainability and that of classification. This model will take a sentence as input and return on the one hand a predicted attention which is considered as the explanation and on the other hand the predicted label which is considered as the classification. Then from the ground truth of attention and label, it calculates the total error which is the sum of the error of attention (explainability) and that of the label (classification) $E_{total} = \lambda E_{ att} + E_{pred}$ \footnote{This loss error was proposed by Mathew in \cite{1}.} and a backpropagation of the total error is made on the model. However, the drawbacks of current HateXplain-based models provide attention that varies considerably when it should be constant. \footnote{This is visible in figure \ref{fig:1}, in the interval [8, 25]}. In some sentences, consecutive words should have similar attention weights due to their shared semantic context. However, attention mechanisms can assign very different weights to these words, resulting in multiple challenges. These include \textbf{inconsistent interpretation} (attention weights not accurately reflecting true word importance), \textbf{prediction instability} (unjustified weight variations affecting output consistency), and \textbf{learning difficulties} (model struggling to converge optimally due to these fluctuations). This leads us to take into account the sequential aspect of the input data when predicting the explanation.

\section{Methodology : Take account a sequential aspect of data on HateXplain model}
\begin{figure}[H]
    \centering
    \includegraphics[width=8.2cm]{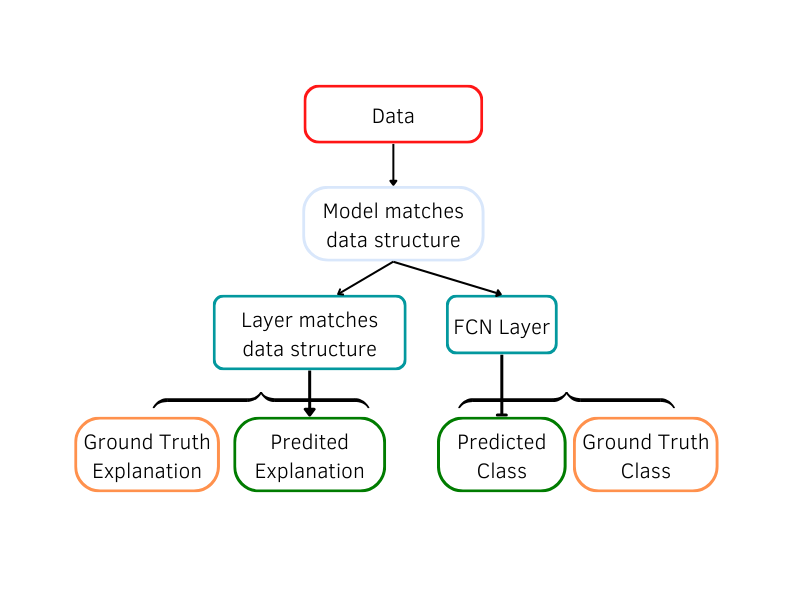}
    \caption{Representation of the general architecture of the model.}
    \label{fig:3.1}
\end{figure}
The idea of this proposal comes from Xu article \cite{19} which proposes to use a deconvolution model to predict an explanation of the places in an image sequence where there has been a movement. We focus more specifically on deconvolution which is compatible with the structure of image data and therefore, it is more able to extract relevant features. We therefore propose a general architecture starting from Xu idea in \cite{19} and Mathew's idea in \cite{1}, by combining Mathew multitask classification and explainability architecture, and the use of a layer compatible with the structure of the input data. 

Figure \ref{fig:3.1} presents this architecture where as input we have any data, then we pass this data to a basic model capable of processing this data perfectly. Then we pass the knowledge extracted from this basic model to a layer compatible with the input data for the prediction of the explanation, and on the other hand we pass this same knowledge to the FCN layer for classification. The idea here is that, if we manage to improve the prediction of the explanation, we can also improve the classification because the learning is multitask and consequently the task of classification and explainability depend on each other. The idea of this proposal comes from Xu article in \cite{19} which proposes to use a deconvolution model to predict an explanation of the places in an image sequence where there has been a movement. We focus more specifically on deconvolution which is compatible with the structure of image data and therefore, it is more able to extract relevant features. 

We therefore propose a general architecture starting from Xu idea in \cite{19} and Mathew idea in \cite{1}, by combining multitask classification and explainability architecture, and the use of a layer compatible with the structure of the input data. Figure \ref{fig:3.1} presents this architecture where as input we have any data, then we pass this data to a basic model capable of processing this data perfectly. We then pass the knowledge extracted from this base model to a layer compatible with the input data for explanation prediction, and then pass this same knowledge to the FCN layer for classification. The idea here is that if we can improve explanation prediction, we can also improve classification because learning is multitask, and consequently, the classification and explainability tasks are interdependent. In the case of graph-type data, for example, we propose the use of GNN (graph neural network: which models the links between the nodes of the graph) for estimating the explanation, which, combined with the multitask approach, will allow for better results. In our case, the data are texts, therefore sequences. We propose a model adapted to our problem of detecting hate speech in Figure \ref{fig:3}. 

The hypothesis we pose here is whether taking into account the contribution dependencies between the elements of the input data by using a layer appropriate to the structure of the input data would allow for a better predicted explanation and consequently a better classification through multitask learning \footnote{During multitask learning, improving one task (explainability)\\ can allow for the improvement of the other task (classification)}.

We start from the fact that in the BiRNN-based HateXplain model called BiRNN-HateXplain \footnote{This model is proposed in this paper : \url{https://arxiv.org/abs/2012.10289}}, we observe a large variation in the attention predicted by the models when this attention should be constant. This shows that the BiRNN-HateXplain model does not take into account the attention dependence on words when predicting it. This is because the model used to calculate attention in the HateXplain models is a \textit{matrix calculation model} which was not designed to take into account precedence dependencies between words in a sentence.
\end{multicols}
\begin{figure}[H]
    \centering
    \includegraphics[width=14cm]{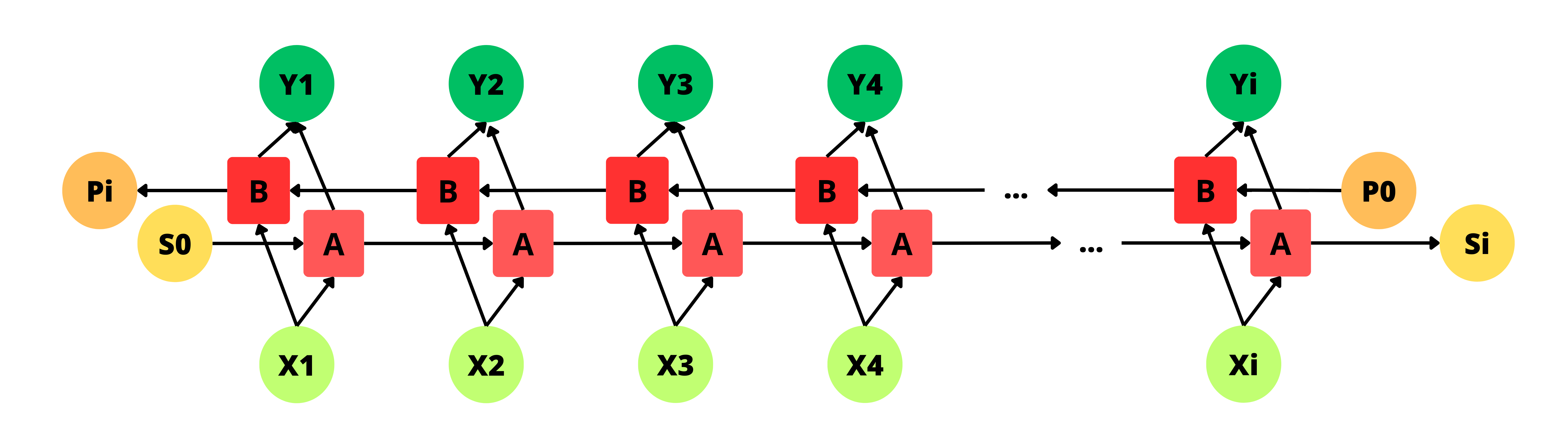}
    \caption{Representation of BiRNN.}
    \label{fig:49}
\end{figure}
\begin{multicols}{2}

\subsection{Bidirectional Recurrent Neural Network (BiRNN)}
Recurrent Neural Networks (RNN) are created with the aim of processing sequential data. RNNs treat data as a sequence of vectors where each vector is processed based on the hidden state of the previous phase, rather than as feed-forward neural networks. The RNN can store data from previous steps of the sequence in some type of memory by calculating the hidden state taking into account both the current input and the hidden state of the previous phase. RNNs are therefore well suited to taking into account the existing dependencies between the elements of the sequence. Many versions of RNN, such as LSTM and GRU, have been proposed to overcome the problem of \textit{vanishing gradients}.

An architecture of a neural network called bidirectional recurrent neural network (BiRNN) is designed to process sequential data. In order for the network to use information from past and future context in its predictions, BiRNNs process input sequences in a forward and backward direction. This is the main distinction between BiRNN and conventional recurrent neural networks which processes the input sequence in one direction.

A BiRNN has two distinct recurrent hidden layers A and B, one of which processes the input sequence forward and the other processes it backward. After that, the results $\{Y_1, Y_2, ..., Y_i\}$ from these hidden layers are collected and inputted into a final prediction layer. As presented in figure \ref{fig:49}, the BiRNN works in the same way as conventional recurrent neural networks, whose recurrent layer $A$, goes in the forward direction $\{X_1 \longrightarrow X_2 \longrightarrow . .., X_{i-1} \longrightarrow X_i\}$, updating the hidden state based on the current input and the previous hidden state at each time step. The backward hidden layer $B$, on the other hand, parses the input sequence in the opposite way $\{X_1 \longleftarrow X_2 \longleftarrow ..., X_{i-1} \longleftarrow X_i\}$, updating the hidden state based on the current input and the hidden state of the next time step. And therefore, the BiRNN takes into account the forward and backward dependencies of elements in a sequence. We choose a BiRNN layer because it is compatible with processing sequence-type data and therefore reflects well the idea proposed above.

\subsection{BiAtt-HateXplain(Bidirectional Attention HateXplain)}
\begin{figure}[H]
    \centering
    \includegraphics[width=8.2cm]{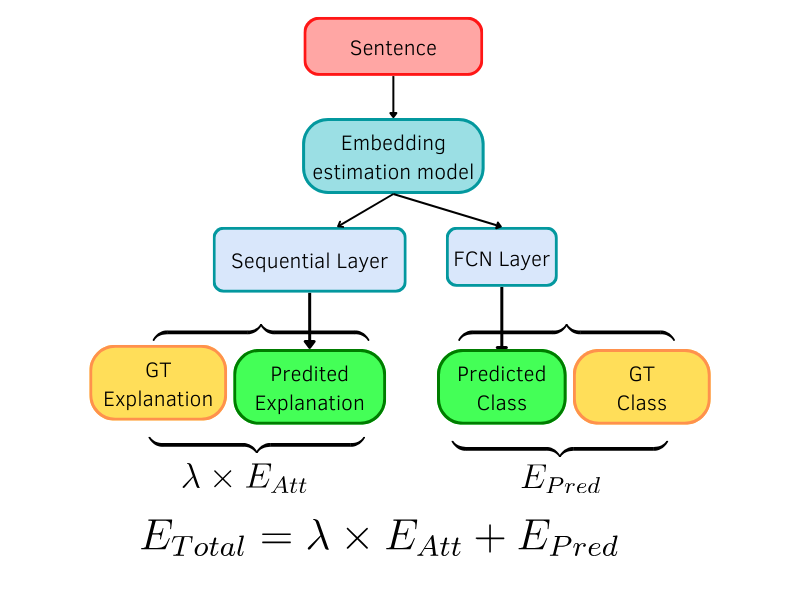}
    \caption{Representation of the general architecture of the model showing the taking into account of the sequential aspect of the data when predicting attention.}
    \label{fig:3}
\end{figure}
We propose to take into account the existing precedence dependencies between the attentions of each word by using a model capable of taking into account this attention dependency when predicting the explanation in HateXplain. More precisely, we propose to use the BiRNN which is a Deeplearning model which was designed to take into account the precedence dependencies between words from left to right and from right to left. This allows us to take into consideration when predicting the explanation attentional dependencies between words in the sentence from left to right as well as from right to left. The model resulting from this consideration of the sequential aspect in the HateXplain models will be the BiAtt-HateXplain (Bidirectional Attention HateXplain) model (figure \ref{fig:3}). 

The hypothesis we pose here is whether \textit{taking into account the contribution dependencies of precedents of the words in the sentence using BiRNN would allow us to solve or mitigate the problem of variations in attention when these attentions should be constant and consequently improve explainability, which will improve classification because the model training is multi-task (dependency between the classification task and the explainability task)}.
\begin{figure}[H]
    \centering
    \includegraphics[width=8.2cm]{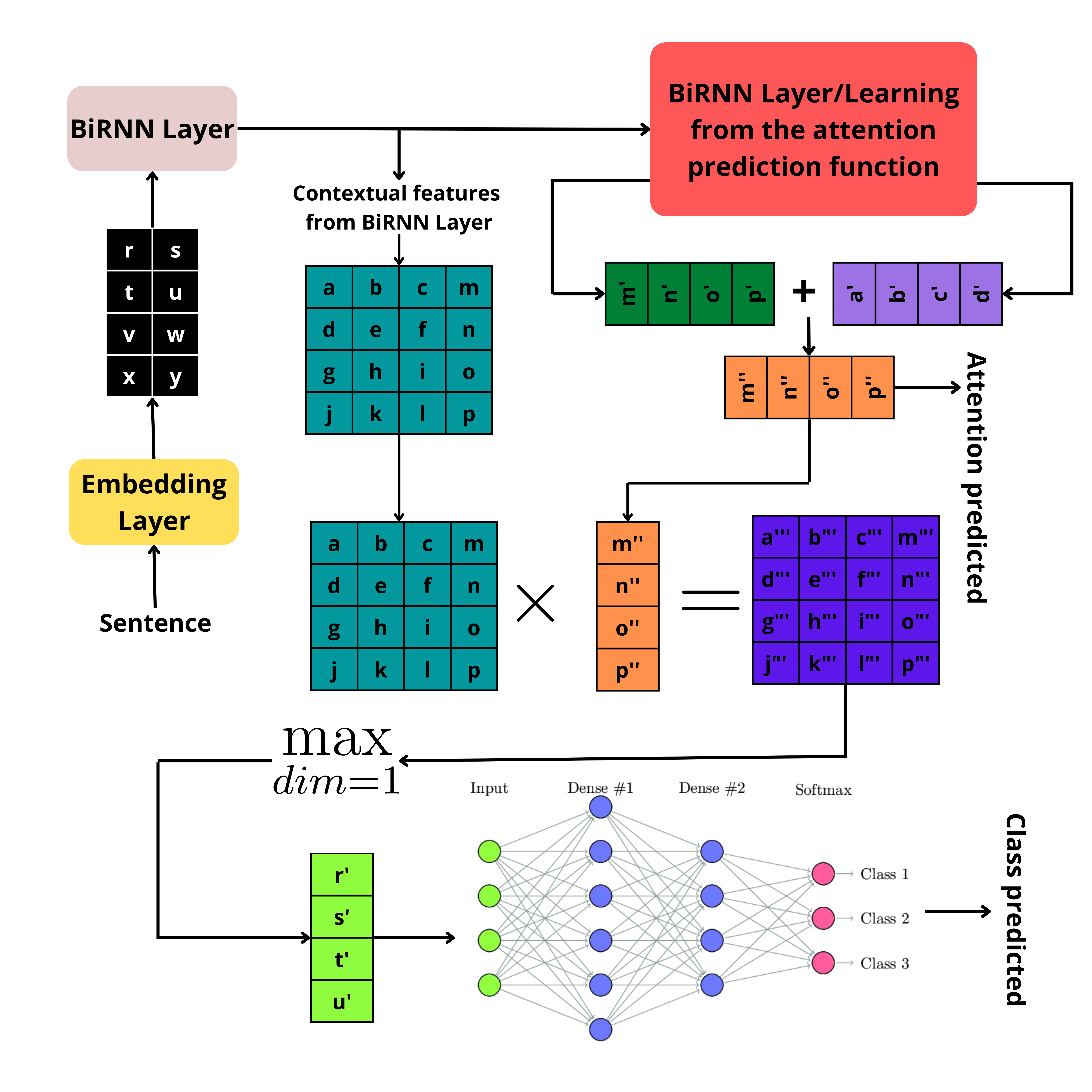}
    \caption{Architecture of the BiAtt-BiRNN-HateXplain model}
    \label{fig:4}
\end{figure}

\subsection{BiAtt-BiRNN-HateXplain(Bidirection BiRNN Hatexplain)}
The BiAtt-BiRNN-HateXplain model is a Deep Learning model based on multi-task learning, classification and explainability. The difference between the two models BiRNN-HateXplain and BiAtt-BiRNN-HateXplain lies at the attention layer. The attention layer of the BiRNN-HateXplain model takes as input the context extracted from the BiRNN layer, then the contextual features of each word in the sentence are used to calculate the attention of each word in the sentence by a process of \textit {matrix multiplication} of these by \textit{learnable tensors} (these tensors are optimized during training of the model). 

On the other hand, the attention layer of the \textit{BiAtt-BiRNN-HateXplain} model (figure \ref{fig:4}) also takes as input the contextual features of each word, but uses a BiRNN layer as an attention learning layer because it is more suitable than the matrix product to capture the left and right attention dependencies of the words in the sentence when learning the \textit{prediction layer attention}. The attention thus predicted can be considered as the explainability of the decision that will be taken by the model. Then, this attention is applied to contextual features to allow the learner to focus attention on certain words in the sentence when making model decisions. An aggregation function $Max$ is then applied to all the words in the sentence by contextual features to have the representation of the features that will be transmitted to the FCN (fully connected neural network) to learn to decide. The decision thus predicted will represent the class assigned to the input sentence by the model. Supervision will also be done in relation to the total error $E_{total} = E_{Pred} + \lambda E_{Att}$.


\section{Experimentations}
In this section, we experiment with the BiAtt-BiRNN-HateXplain algorithm \footnote{\url{https://github.com/pharaon-dev/BiAttention-HateXplain}} with the aim of evaluating these capabilities in terms of performance, bias and explainability. The data used in the experiment are the HateXplain\footnote{\url{https://github.com/hate-alert/HateXplain}} data proposed by \cite{1}. This experiment was carried out on a 16GB RAM using the pytorch\footnote{\url{https://pytorch.org/} or \url{https://github.com/pytorch}} tool for learning the models.

\subsection{HateXplain}
HateXplain of \cite{1} is the first benchmark dataset on hate speech taking into account the ground truth of explainability and contains 20,000 messages. Each publication in this dataset is annotated from 3 perspectives:
\begin{itemize}
    \item The basic classification has three commonly used classes: Hateful, Offensive, Normal \footnote{After the annotation of the messages by three annotators, the truth class is obtained by majority vote.}.
    \item The target community (the one that was the victim of hate speech/offensive speech in the post)\footnote{To decide the target community of a post, we rely on majority voting.}.
    \item Justifications (explainability): The parts of the message on which their decision to label it as hateful or offensive is based. Examples of message annotation in HateXplain are presented in the Table \ref{tab:8}. 
\end{itemize}
After annotating a message by three annotators, the process of calculating the ground truth attention is as follows:
\begin{itemize}
    \item Mathew in \cite{1} proposes to first convert each justification into an \textbf{attention vector}. This is a Boolean vector of length equal to the number of tokens in the sentence. Tokens in the justification are indicated by a value of $1$ in the attention vector.
    \item If a speech is hateful, we take the average of these attention vectors to represent a common base attention vector for each message. It proposes to normalize this common attention vector via a $Softmax$ function to generate \textit{ground truth attention}.
    \item Finally, if the post label is $normal$, we ignore the attention vectors and replace each element of the ground truth attention with \textit{1/(sentence length)} to represent a uniform distribution.
\end{itemize}
\end{multicols}
\begin{table}[H]
\centering
\caption{Examples from the HateXplain dataset. The \colorbox{green}{highlighted} part of the text represents the annotator's justification.} 
\label{tab:8}
\resizebox{1.0\columnwidth}{!}{%
\begin{tabular}{l|l}
\hline
\\
Text & Dad should have told the \colorbox{green}{muzrat whore}
to \colorbox{green}{fuck off}, and went in anyway\\
Label & Hate\\
Targets & Islam \\
\hline
\\
Text & A \colorbox{green}{nigress too dumb to fuck} has a scant
chance of understanding anything beyond
the size of a dick \\
Label & Hate \\
Targets &  Women, African \\
\hline
\end{tabular}%
}
\end{table}

\begin{multicols}{2}
\subsection{Evaluation Metrics}
The metrics used here are even used by Mathew \cite{1}. We have metrics based on performance, bias and explainability. 
As performance-based metrics, we have accuracy, F1 macro score, and AUROC score. As Bias metrics we have GMB-Subgroup-AUC, GMB-BPSN-AUC and GMB-BNSP-AUC. As an explainability metric we have the plausibility and reliability metrics. These metrics help assess whether a toxic classification model is biased or fair toward certain communities mentioned in comments. They measure how accurately the model handles comments related to sensitive groups: 
\begin{itemize}
    \item Subgroup-AUC measures the model's ability to distinguish toxic from normal comments within a particular group, using only posts (toxic and normal) that mention that group, and therefore, the higher the value, the better the model distinguishes within that group.
    \item BPSN-AUC (Background Positive, Subgroup Negative) Measures the model's tendency to confuse a normal comment mentioning a group with a toxic comment that does not mention it.
    \item BPSN-AUC (Background Positive, Subgroup Negative) Measures the model's tendency to confuse a normal comment mentioning a group with a toxic comment that doesn't. Therefore, the higher the value, the fewer false positives related to that group.
    \item GMB-AUC (Generalized Mean of Bias AUCs) provides an overall measure of the model's bias toward all identified groups by taking a generalized average of the Subgroup-AUC, BPSN-AUC, and BNSP-AUC metrics; we therefore obtain the following metrics: GMB-Subgroup-AUC, GMB-BPSN-AUC, and GMB-BNSP-AUC.
\end{itemize}
As explainability based metrics, we have : 
\begin{itemize}
    \item Faithfulness : Measures the extent to which the explanation truly reflects the model's internal reasoning, rather than simply "making sense" to a human. Metrics used:
    \begin{itemize}
        \item Comprehensiveness: Measures how much the model's output decreases when tokens considered important are removed. If performance drops significantly, the explanation is considered faithful.
        \item Sufficiency: Measures how well the output remains correct when only the tokens considered important are retained. If performance remains high, this shows that these tokens are sufficient for the model's reasoning.
    \end{itemize}
    \item Plausibility : Measures how credible or convincing an explanation seems to a human.
    Metrics for discrete token selection:
    \begin{itemize}
        \item IOU F1-Score: Intersection over Union (IoU) compares a set of predicted tokens and human annotations (rationales): An overlap greater than 0.5 with one of the human justifications is considered a match. Allows the calculation of an F1 score based on partial matches.
        \item Token F1-Score: Based on the precision and recall of predicted tokens vs. human rationales. Calculates the F1-score directly on the selected tokens.
        \item AUPRC (Area Under the Precision-Recall Curve): Used when tokens have continuous scores (rather than being selected in a binary fashion). Measures the quality of the scores assigned to tokens when applying different thresholds.
    \end{itemize}
\end{itemize}

\subsection{Hyper-parameter tuning}
The BiRNN \cite{12}, BiRNN-Attention \cite{11} and CNN-CRU \cite{21} models have hyper-parameter settings similar to that of Mathew \cite{1}. On the other hand, the BiAtt-BiRNN models will use Adam as optimizer and $0.001$ as learning rate. Optimal performance of BiAtt BiRNN occurs with $\lambda$ being set to 100. We also use the pre-trained crawl GloVe embeddings \cite{17} to initialize word embeddings on the BiAtt-BiRNN model. In the case of BiAtt-BiRNN a dropout layer was added before the BiRNN layer used for explainability prediction, we tested dropout values from $0.1$ to $0.5$, the dropout with a value $0.2$ on the model BiAtt-BiRNN-HateXplain allowed us to have better results. The table \ref{tab:1} presents the different variations of hyper-parameters tested, those of the state-of-the-art models had been proposed in \cite{1}.
\end{multicols}
\begin{figure}
    \caption{Attention predicted by BiRNN-HateXplain in \textcolor{yellow}{\textit{yellow}}, Attention predicted by BiAtt-BiRNN-HateXplain in \textcolor{red}{\textit{red}}, and ground truth attention in \textcolor{green}{\textit{green}}}
    \label{fig:5}
    \includegraphics[width=.53\textwidth]{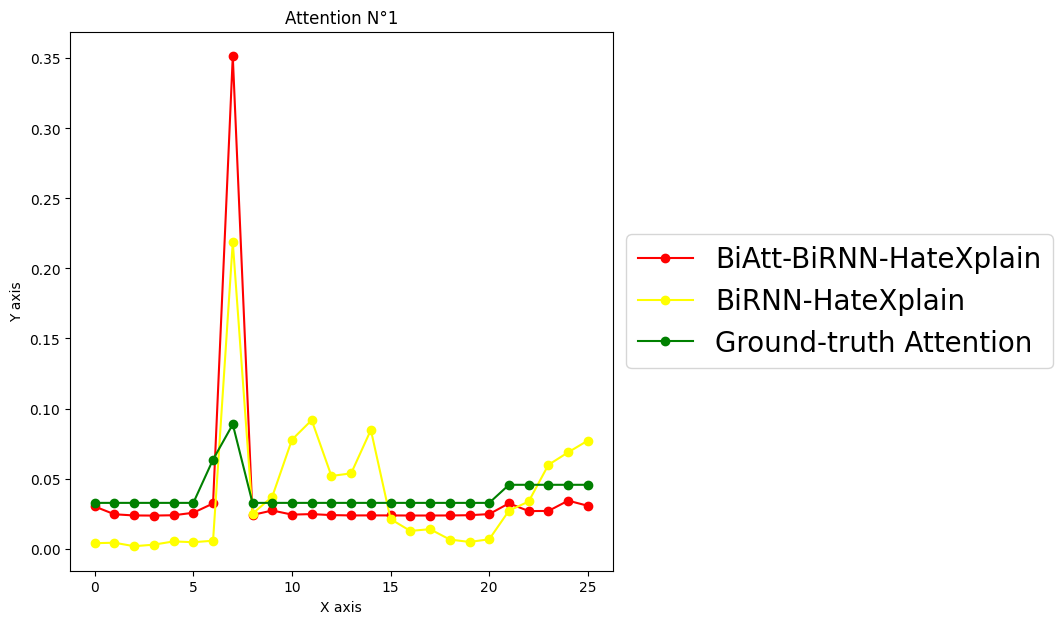}\hfill
    \includegraphics[width=.53\textwidth]{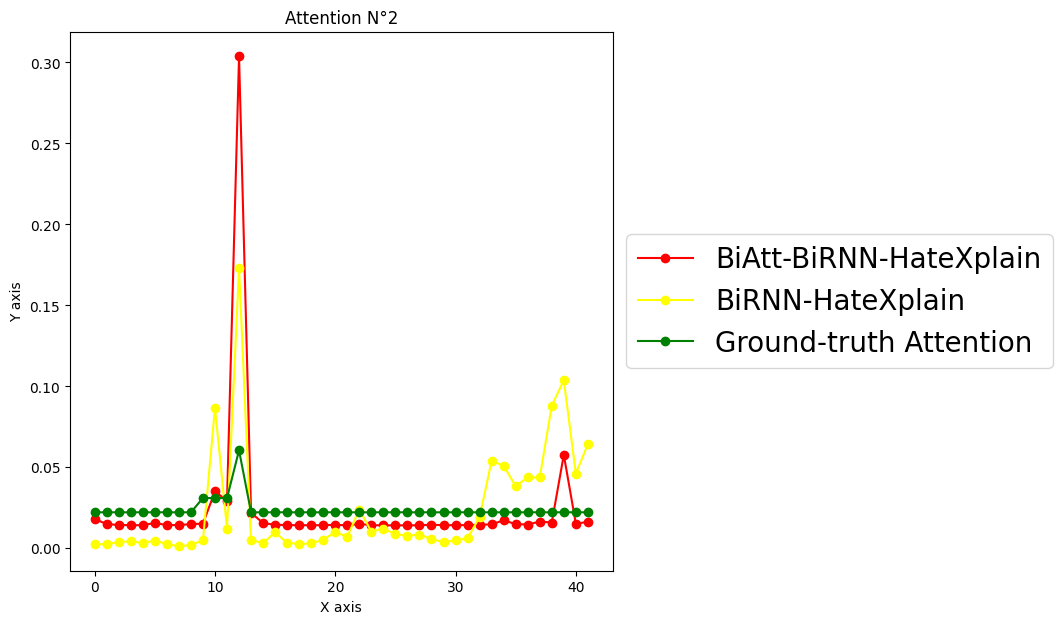}\hfill
    \\[\smallskipamount]
    \includegraphics[width=.53\textwidth]{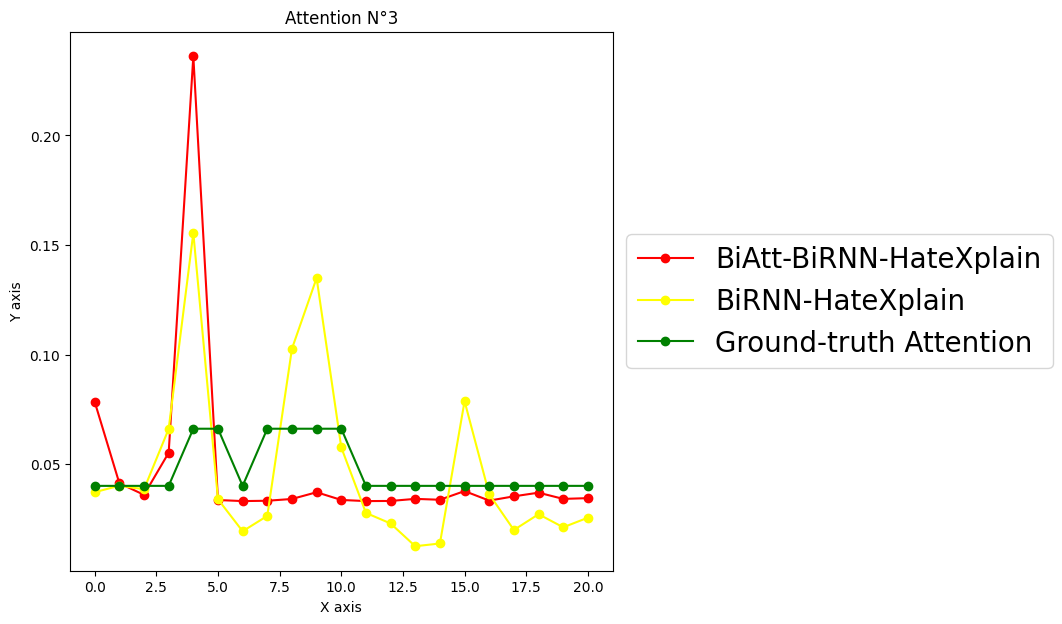}\hfill
    \includegraphics[width=.53\textwidth]{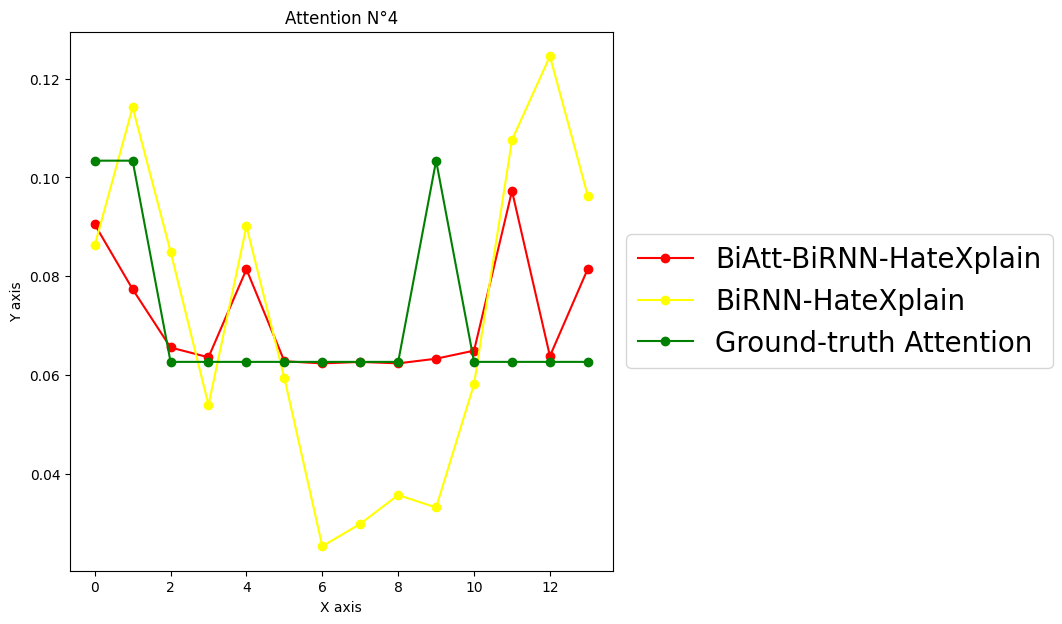}\hfill
    \\[\smallskipamount]
    \includegraphics[width=.53\textwidth]{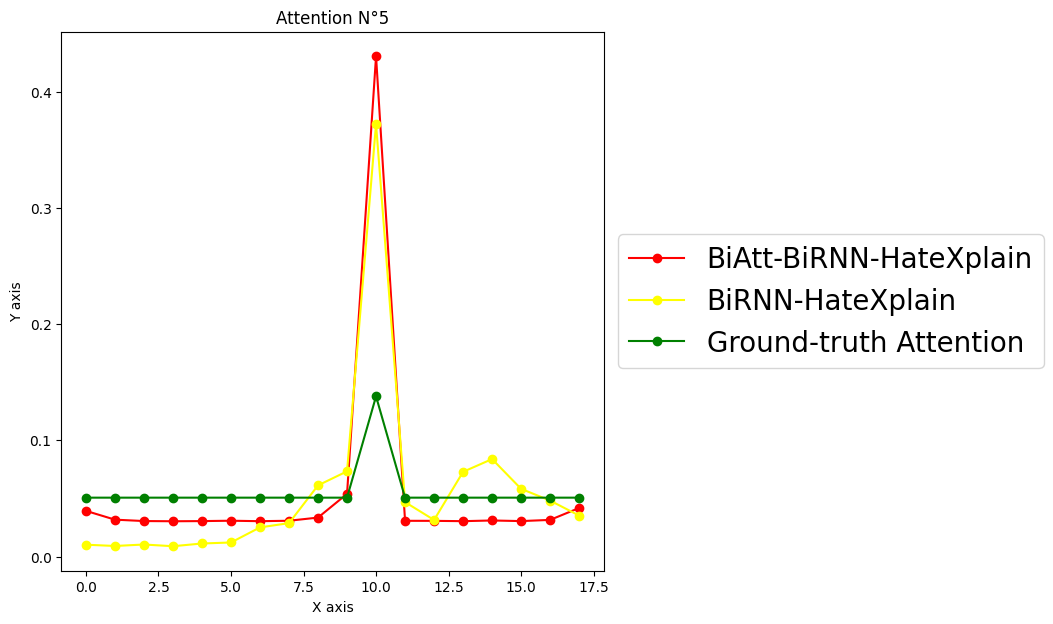}\hfill
    \includegraphics[width=.53\textwidth]{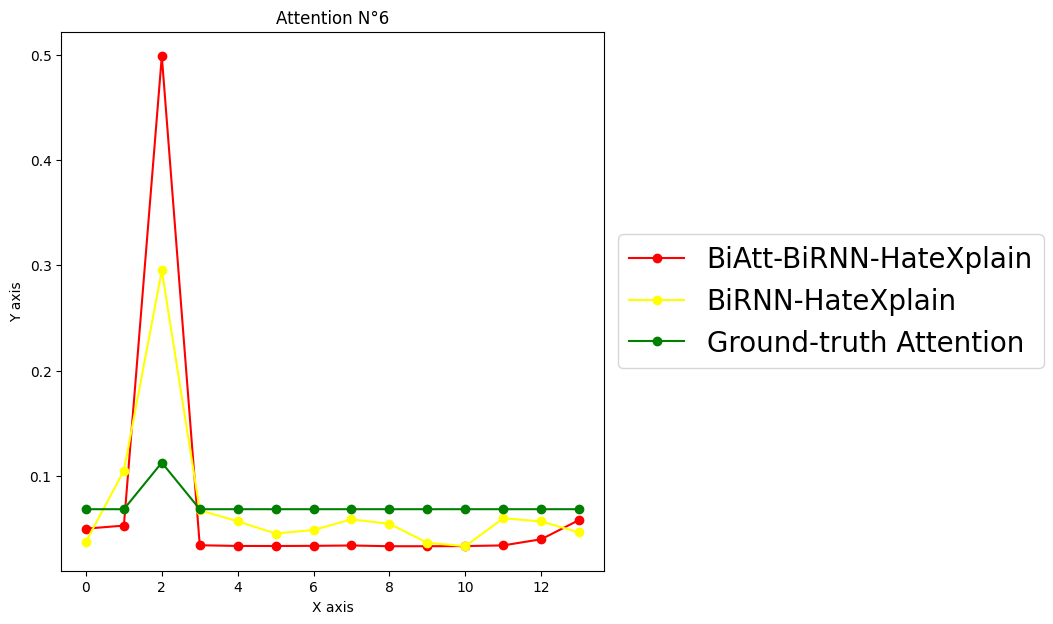}\hfill
    \\[\smallskipamount]
    \includegraphics[width=.53\textwidth]{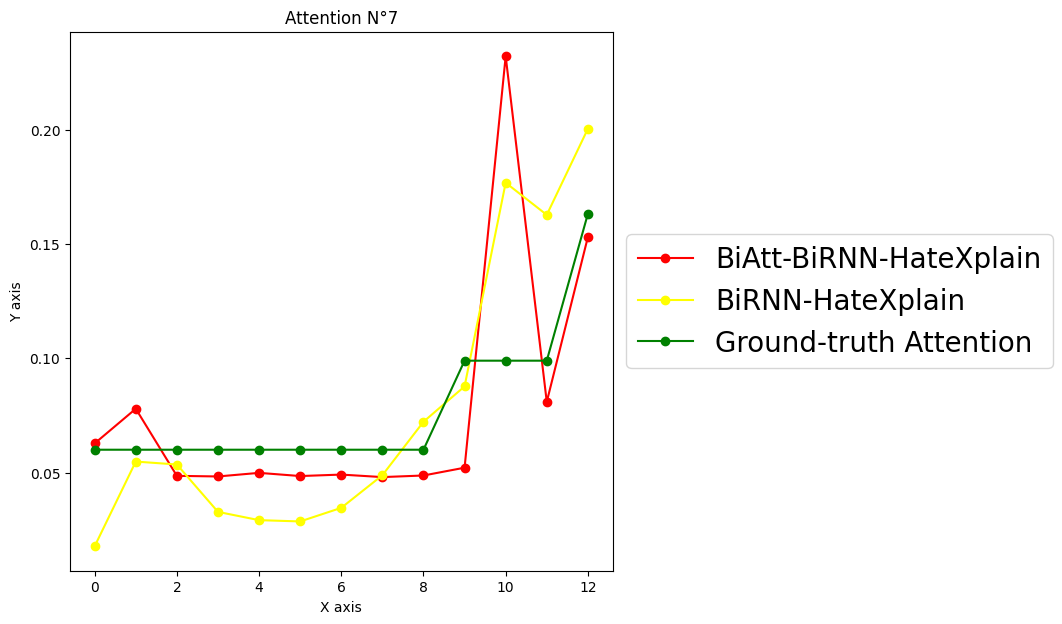}\hfill
    \includegraphics[width=.53\textwidth]{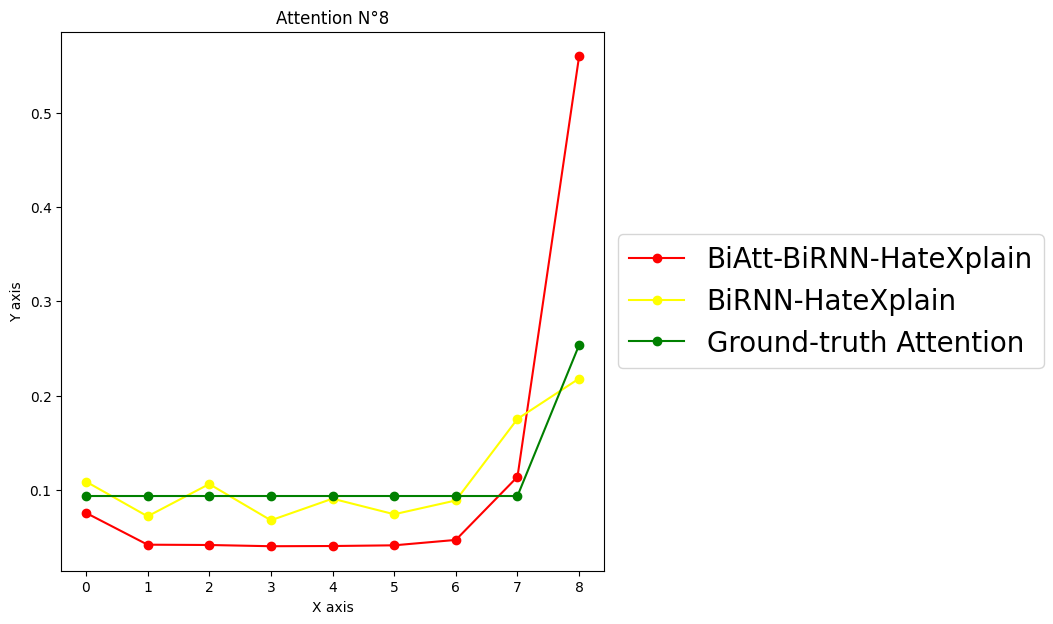}\hfill
\end{figure}

\begin{table}[!ht]
\centering
\caption{The hyper-parameter variations of our model and those of Mathew proposed in [1]} 
\label{tab:1}
\resizebox{1.0\columnwidth}{!}{%
\begin{tabular}{|l|c|c|c|c|c|}
    \hline
    \textbf{Hyper-parameters} & \textbf{BiRNN} & \textbf{BiRNN-Attention} & \textbf{CNN-GRU} & \textbf{BiAtt-BiRNN} \\ 
    \hline
    \textbf{No. of hidden units in sequential layer} & \textbf{64, 128} & \textbf{64, 128} & \textbf{-NA-} & \textbf{64, 128}\\
    \hline
    \textbf{Sequential layers type}& \textbf{LSTM,GRU} & \textbf{LSTM,GRU} & \textbf{GRU} & \textbf{LSTM,GRU}\\
    \hline
    \textbf{Train embedding layer} & \textbf{True, False} & \textbf{True, False} & \textbf{True, False} & \textbf{True, False} \\
    \hline
    \textbf{Dropout after embedding layer} & \textbf{0.1, 0.2, 0.5} & \textbf{0.1, 0.2, 0.5} & \textbf{0.1, 0.2, 0.5} & \textbf{0.1,0.2, 0.3, 0.4, 0.5}\\
    \hline
    \textbf{Dropout after fully connected
    layer} & \textbf{0.1,0.2,0.5} & \textbf{0.1,0.2,0.5} & \textbf{0.1,0.2,0.5} & \textbf{0.1,0.2, 0.3, 0.4, 0.5} \\
    \hline
    \textbf{Dropout before BiRNN layer}& \textbf{-NA-} & \textbf{-NA-} & \textbf{-NA-} & \textbf{0.1 ,0.2, 0.3, 0.4, 0.5} \\
    \hline
    \textbf{Learning rate} & \textbf{0.1,0.01,0.001} & \textbf{0.1,0.01,0.001} & \textbf{0.1,0.01,0.001} & \textbf{0.1,0.01,0.001} \\
    \hline
    \textbf{For supervised part} \\
    \hline
    \textbf{Attention lambda ($\lambda$)}& \textbf{-NA-} & \textbf{0.001,0.01,0.1,1,10,100} & \textbf{-NA-} & \textbf{0.001,0.01,0.1,1,10,100}\\
    \hline
\end{tabular} 
}
\end{table}
\begin{table}[!ht]
\centering
\caption{\textit{Results in terms of performance metrics, bias and explainability}} 
\label{tab:3}
\resizebox{1.0\columnwidth}{!}{%
\begin{tabular}{l|ccc|ccc|ccc|cc}
\multicolumn{1}{l}{Model [Token Method]}&\multicolumn{2}{c}{Performance}&\multicolumn{3}{c}{Biais}&\multicolumn{4}{c}{Explainability}\\
 & & & & & & & Plausibility & & &Faithfulness & \\
 &Acc. &Macro F1 &AUROC & GMB-Sub & GMB-BPSN & GMB-BNSP & IOU F1 &Token F1 & AUPRC & Comp. & Suff.\\
\hline
CNN-GRU[LIME]&0.627 &0.606 &0.793 &0.654 &0.623 & 0.659& 0.167 &0.385 &0.648 &0.316 &-0.082\\
\hline
BiRNN[LIME]&0.595 &0.575 &0.767 &0.640 &0.604 &0.671& 0.162 &0.361 &0.605 &\textbf{0.421} &-0.051\\
\hline
BiRNN-Attn[Attn]&0.621 &0.614 &0.795 &0.653 &0.662 &0.668& 0.167 &0.369 &0.643 &0.278 &0.001\\
\hline
BiRNN-Attn[LIME]&0.621 &0.614 &0.795 &0.653 &0.662 &0.668& 0.162 &0.386 &0.650 &0.308 &-0.075\\
\hline
BiRNN-HateXplain[Attn]&0.629 &0.629 &0.805 &0.691 &0.636 &0.674& 0.222 &\textbf{0.506} &\textbf{0.841} &0.281 &0.039\\
\hline
BiRNN-HateXplain[LIME]&0.629 &0.629 &0.805 &0.691 &0.636 &0.674&0.174 & 0.407 &0.685 &0.343 & -0.075 \\
\hline
\textbf{BiAtt-BiRNN-HateXplain}[Attn]& \textbf{0.65} & \textbf{0.64} & \textbf{0.81} & \textbf{0.734} & \textbf{0.724} & \textbf{0.706}& \textbf{0.230} & 0.487 & 0.823 & 0.310 & -0.060\\
\hline
\textbf{BiAtt-BiRNN-HateXplain}[LIME]& \textbf{0.65} & \textbf{0.64} & \textbf{0.81} & \textbf{0.734} & \textbf{0.724} & \textbf{0.706}& 0.188 & 0.409 & 0.738 & 0.318 & \textbf{-0.126}\\
\hline
\end{tabular}%
}
\end{table}

\begin{multicols}{2}

\section{Results \& Observations}
In figure \ref{fig:5}, we observe on the plot Attention N°{1-8} that the attention predicted by the proposed model BiAtt-BiRNN-HateXplain is constant very often where it should be constant. For example at Attention N°1 we observe that on the interval [8, 20] that the ground truth attention is constant and the attention predicted by the BiRNN-HateXplain model varies a lot, on the other hand, the attention predicted by the proposed model BiAtt-BiRNN-HateXplain is constant on this interval which shows that taking into account the sequential aspect of input data using the BiRNN has made it possible to improve the estimation of attention or explanation in this case. This observation can also be made on other plots such as the Attention plot N°6 which shows that compared to the attention predicted by BiRNN-HateXplain which varies a lot over the interval [3, 10], the attention predicted by the proposed model BiAtt-BiRNN-HateXplain is constant like that of the ground truth. In addition to this, we observe that the attention predicted by the proposed model BiAtt-BiRNN-HateXplain is very often close to that of the ground truth while remaining constant when it should be. This is for example observed on the Attention plot N°1 in the interval [8,20], on the Attention plot N°2 in the interval [14, 41], on the Attention plot N°3 in the interval [11, 20], on the Attention plot N°5 in the interval [1, 8], and on the Attention plot N°7 in the interval [2, 8].

The results in terms of metrics presented on the tables \ref{tab:3} of the BiAtt-BiRNN-HateXplain model show that there is an improvement in explainability in BiAtt-BiRNN-HateXplain[Attn \& LIME], more precisely of \textit{fidelity} when we take into account the sequential aspect of the data when learning the \textit{attention prediction layer}. Which shows that these models focus much more on \textit{words with high attention} to decide. We observe an improvement in \textbf{IOU F1} of the BiAtt-BiRNN HateXplain[Attn] model, which shows that the model increasingly predicts attentions \textit{similar} to that of the ground truth.

Also note, for the proposed model, a decrease in the values of the \textit{plausibility} metrics: \textit{Token F1}, \textit{AUPRC}. On the other hand, we observe an increase in all \textit{plausibility} metrics of the BiAtt-BiRNN-HateXplain[LIME] model compared to the BiRNN-HateXplain[LIME] model, which shows that the explainability of this model with LIME is increasingly convincing for humans. We see an increase in the performance of the BiAtt-BiRNN-HateXplain model compared to the previous BiRNN-HateXplain model, which shows that these models are learning better and better to detect hate speech. We also observe a large increase in Bias metrics on the BiAtt-BiRNN-HateXplain model, compared to BiRNN-HateXplain, which shows that the decisions of the BiAtt-BiRNN-HateXplain model are less and less influenced by \textit{unintentional biases}.
\begin{table}[H]
\centering
\caption{\textit{Another example}} 
\label{tab:4}
\resizebox{1.0\columnwidth}{!}{%
\begin{tabular}{l|c|c} \hline
    {$Model$} & {$Text$} & {$Label$} \\ \hline
    Human Annotator  & \colorbox{green}{I} \colorbox{green}{hate} \colorbox{green}{arabs} & \textbf{HS} \\ \hline
    BERT  & I \colorbox{green}{hate} arabs & \textbf{Normal} \\
    BERT-HateXplain  & I \colorbox{green}{hate} \colorbox{green}{arabs} & \textbf{HS} \\
    BiAtt-BiRNN-HateXplain  &\colorbox{green}{I} \colorbox{green}{hate} \colorbox{green}{arabs}  & \textbf{OF} \\ \hline
\end{tabular}
}
\end{table}
Table \ref{tab:4} shows that using explainability by attention on the example presented, the BiAtt-BiRNN-HateXplain model explains better than BERT-HateXplain, but does not classify better than the latter.

\section{Conclusion}
In this work, our objective was to improve the explainability of artificial neural networks, considered as black boxes, by taking into account the sequential aspect of the input data to solve the problem of variability of the predicted attention when it should be constant. To address this concern, we proposed to use the BiRNN model for training the explainability prediction layer rather than the BiRNN-HateXplain matrix approach and obtained the BiAtt-BiRNN-HateXplain (Bidirectional Attention) model. During experiments on the HateXplain benchmark, we found that taking into account the sequential aspect of the data improved the performance, biases, but especially the explainability of the model. In future work, we plan to integrate the concept of considering the sequential aspect of data when estimating explainability into the BERT-HateXplain model or into a model like LIME.

\end{multicols}
\end{document}